%% file: main.tex
\pdfoutput=1

\documentclass[11pt]{article}

\usepackage[preprint]{acl}

\usepackage{times}
\usepackage{latexsym}

\usepackage[T1]{fontenc}

\usepackage[utf8]{inputenc}

\usepackage{microtype}

\usepackage{inconsolata}

\usepackage{graphicx}
\usepackage{amsmath}
\usepackage{hyperref}
\usepackage{booktabs}
\usepackage{algorithm}
\usepackage{algorithmic}
\usepackage[switch]{lineno}
\usepackage{graphicx}
\usepackage{amssymb}
\usepackage{multirow}
\usepackage{pifont}
\usepackage{arydshln} 
\newcommand{\cmark}{\ding{51}}%
\newcommand{\xmark}{\ding{55}}%
\title{On the Evolution of Federated Post-Training Large Language Models:\\ A Model Accessibility View}

\author{
  Tao Guo \\
  Shenzhen University\\
  \And
  Junxiao Wang \\
  Guangzhou University\\
  \And
  Fushuo Huo \\
  PolyU \\
  \And
  Laizhong Cui \\
  Shenzhen University \\
  \AND
  Song Guo \\
  HKUST \\
  \And
  Jie Gui \\
  Southeast University \\
  \And
  Dacheng Tao \\
  NTU \\
}

\begin{document}
\maketitle
\begin{abstract}

Federated Learning (FL) enables training models across decentralized data silos while preserving client data privacy. Recent research has explored efficient methods for post-training large language models (LLMs) within FL to address computational and communication challenges.
While existing approaches often rely on access to LLMs' internal information, which is frequently restricted in real-world scenarios, an inference-only paradigm (black-box FedLLM) has emerged to address these limitations.
This paper presents a comprehensive survey on federated tuning for LLMs. We propose a taxonomy categorizing existing studies along two axes: model access-based and parameter efficiency-based optimization. We classify FedLLM approaches into white-box, gray-box, and black-box techniques, highlighting representative methods within each category. We review emerging research treating LLMs as black-box inference APIs and discuss promising directions and open challenges for future research.

\end{abstract}

\begin{figure}[t]
  \centering  \includegraphics[width=1.0\linewidth]{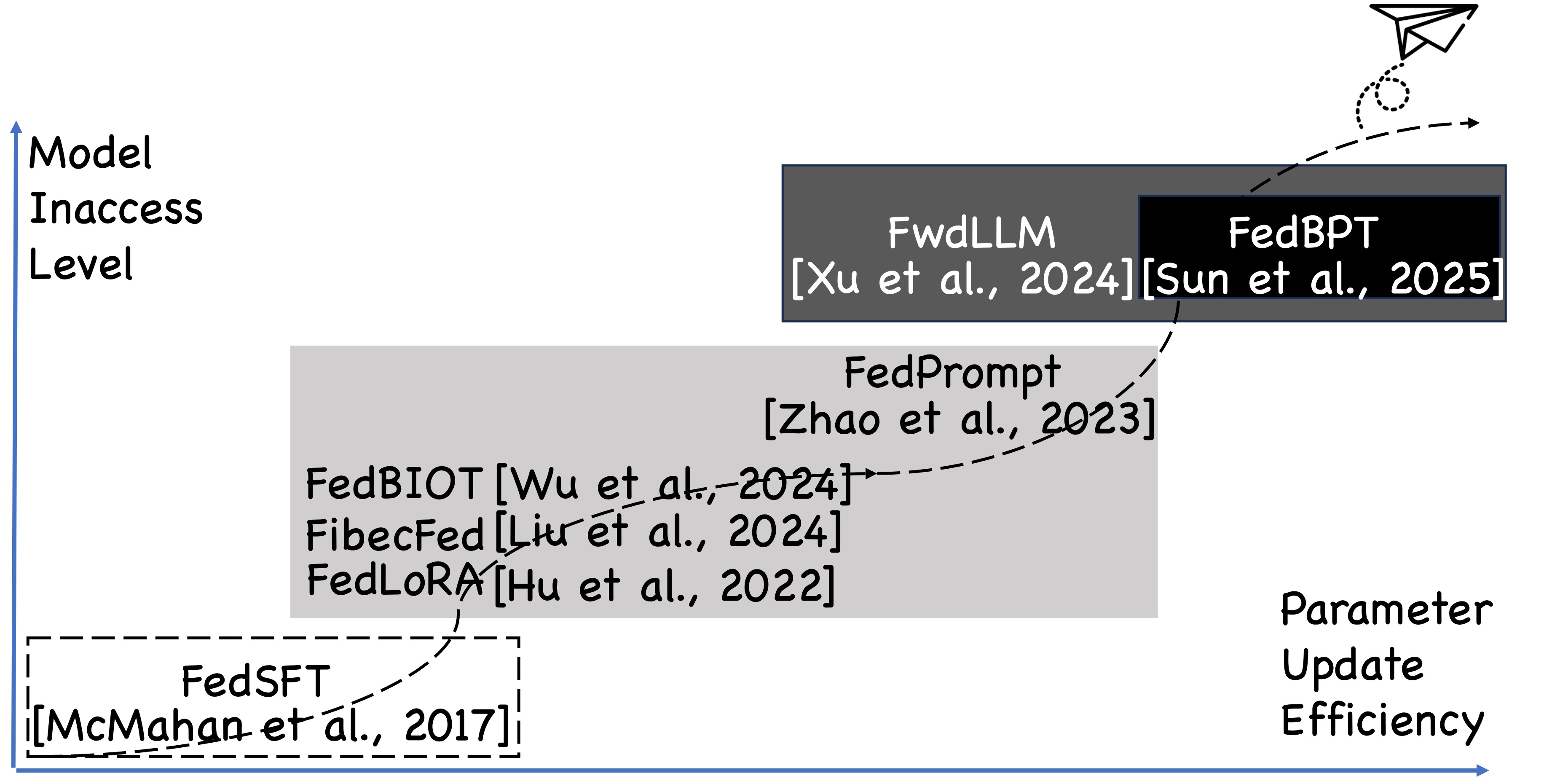}

  \caption{Evolution path of FedLLM. We show the evolution path as well as the representative studies along FedLLM development. Researchers first combine Fedavg with LLM  to mutually benefit both paradigms, as shown in the white area. As the field progressed, the focus shifted toward improving efficiency to mitigate the substantial computational and communication overhead introduced by LLMs, as shown in the gray area. More recently, increasing attention has been directed toward optimizing both integration access and efficiency-based constraints, as shown in the black area. }

  \label{fig:evolution}
\end{figure}

\section{Introduction}
Large Language Models (LLMs) have showcased remarkable capabilities. As foundational models, pre-trained LLMs can be fine-tuned for a wide range of downstream tasks and applied across various domains. However, the post-training process often involves sensitive user data, raising significant privacy and security concerns. Federated Learning (FL) \cite{mcmahan2017communication} offers a decentralized, collaborative approach to fine-tuning LLMs while maintaining data minimization, a method commonly referred to as FedLLM.

Deploying FL for LLMs poses significant challenges, including high computational demands on local clients and excessive communication overhead, which can hinder practical implementation. To address these challenges, recent research has explored various efficient FedLLM methods, commonly referred to as parameter-efficient fine-tuning techniques. One notable advancement in computational and communication efficiency is the use of low-rank adapters (LoRA) \cite{hu2022lora}. LoRA reduces the dimensionality of weight update matrices in transformer modules, achieving a lower-dimensional representation without the need for additional adapters. FedLoRA applies this technique in FL systems by building adapters for existing parameters across all clients, improving efficiency without altering the model architecture. Another parameter-efficient fine-tuning method, prompt tuning \cite{lester2021power}, adapts pre-trained LLMs to downstream tasks without training the model itself. Instead, it relies on designing textual prompts that are prepended to input queries. This approach keeps the model entirely frozen, modifying only the prompts (a small set of tokens) to enhance output quality. Unlike vanilla fine-tuning, prompt tuning avoids altering the model's parameters or architecture. Both FedLoRA and FedPrompt usually achieve performance comparable to (not significantly less than) full-model fine-tuning in FL, demonstrating their potential as efficient and practical solutions for federated post-training of LLMs.

However, after thoroughly examining existing surveys on federated post-training for LLMs \cite{ijcai2024p919, chen2024integration, yao2024federated, yan2025federated}, we identified a significant gap in the literature: Few comprehensive studies address the challenges posed by restricted access to model parameters, particularly in inference-only and black-box settings. In summary, this paper makes the following key contributions.

\begin{itemize}
    \item 
    We introduce a novel FedLLM taxonomy along two axes: model access-based and parameter efficiency-based optimization. The access-based approach addresses federated tuning in inference-only and black-box settings, while the efficiency-based approach tackles high computational demands and communication overhead. Using these axes, we classify FedLLM approaches into white-box, gray-box, and black-box techniques, detailing representative methods for each.

    \item Next, we review emerging and representative studies that treat LLMs as black-box, focusing primarily on gradient-free optimization techniques, and emphasize the significance of the inference-only API approach. 
    By analyzing pioneering work that combines access-based and efficiency-based optimization, we trace FedLLM's evolutionary trajectory toward achieving efficiency in inference-only, black-box settings.

    \item 
    We conclude by identifying open challenges and future research directions, focusing on federated value alignment in inference-only and black-box settings through DPO, RLHF, and RLAIF approaches. Additionally, we emphasize the critical need for enhanced security and privacy measures in these constrained environments.

    \item To the best of our knowledge, this is the first comprehensive survey to explore the broad integration of model access-based and parameter efficiency-based optimization in FedLLM.
\end{itemize}

\begin{figure*}[t]
  \centering  \includegraphics[width=0.9\linewidth]{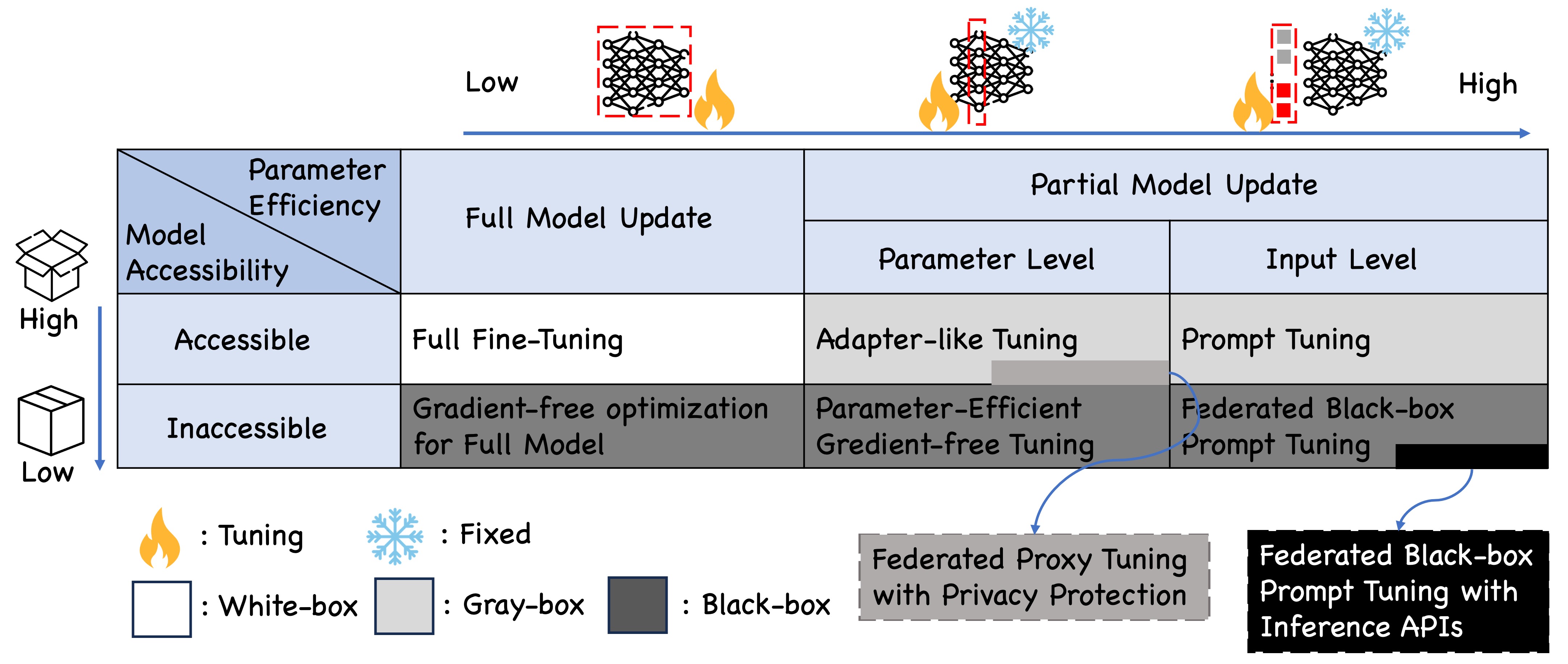}
  \vspace{-0.3cm}
  \caption{Taxonomy of Federated LLM. We re-examine the existing Federated LLM approaches from the `Model Accessibility' and `Parameter Efficiency' dimensions and categorize the existing Federated LLM methods into the above three types. Specifically, the white box represents white-box tuning, the gray box represents gray-box tuning and the black box represents black-box tuning. Specifically, the absolute black area denotes the federated black-box prompt tuning with inference APIs, achieving the highest parameter efficiency and privacy protection. }
  \vspace{-0.3cm}
  \label{fig:framework}
\end{figure*}

\input{sections/Taxonomy}

\input{sections/Whitebox}
\input{sections/Graybox}
\input{sections/Blackbox}

\section{Discussions \& Future Directions}
After reviewing the existing research on FedLLM, we identify several unresolved issues that warrant further exploration. While current studies primarily focus on improving the efficiency of the supervised fine-tuning (SFT) stage in inference-only and black-box settings, the crucial stage of federated value alignment remains underexplored. To this end, we highlight several promising directions for advancing the federated value alignment stage, such as direct preference optimization (DPO), reinforcement learning from human feedback (RLHF), and reinforcement learning from AI feedback (RLAIF).

\noindent\textbf{Federated DPO.} 
DPO aligns LLMs by training on pairs of preferred and dis-preferred responses \cite{rafailov2023direct}. In the FedLLM context, human values can be integrated into LLMs and further enriched by involving a large number of clients, thereby capturing a diverse range of human values. Unlike other RL-based methods that require a reward model, Federated DPO (FedDPO) serves as a practical approach for federated value alignment by collaboratively fine-tuning SFT models using local preference data. Therefore, applying existing efficiency-based optimization methods to FedDPO and integrating them into a comprehensive workflow of FedLLM represents a straightforward future direction.

\noindent\textbf{Federated RLHF.} 
Compared to FedDPO, Federated RLHF (FedRLHF) is relatively more complex due to its two-stage process requiring a reward model before fine-tuning the LLM itself \cite{bai2022training}, which can be tedious in federated settings. Additionally, FedRLHF tends to be more unstable and may struggle to converge, especially when facing the inherent statistical heterogeneity in FL systems, making a straightforward implementation of FedRLHF impractical. However, recent advances in centralized methods like Proximal Policy Optimization (PPO) incorporate optimizations to address policy gradient instability. These countermeasures offer promising solutions to produce a more stable and efficient FedRLHF, making it a worthwhile direction for future exploration.

\noindent\textbf{Federated RLAIF.} RLHF has been highly effective in aligning LLMs with human preferences. However, gathering high-quality preference labels is costly, particularly in FL systems. The recently proposed RLAIF framework offers a promising alternative by leveraging preferences generated by the LLM itself to train the reward model \cite{bai2022constitutional}. Experimental results indicate that RLAIF can achieve performance comparable to RLHF, presenting a potential solution to the scalability challenges associated with RLHF \cite{lee24rlaif}. Nonetheless, RLAIF annotates preferences using an off-the-shelf LLM—a model that is pre-trained or instruction-tuned for general purposes but not fine-tuned for specific downstream tasks. This raises a question: can the pre-trained model in FedLLM effectively generate preferences to reward its fine-tuned version and create a self-contained FL system? If not, the reliance on additional LLMs as auxiliary tools for local clients in FedLLM could further complicate these systems. Addressing these compelling questions remains an exciting avenue for future research.

\noindent\textbf{Privacy and Security.} Treating LLMs as black-box inference APIs helps bridge the gap when model access is restricted, but also introduces new concerns regarding privacy and security. On the one hand, users may need to query the model APIs with their private inputs, necessitating the use of lightweight secure inference techniques to safeguard privacy. On the other hand, the black-box nature of these APIs makes it more challenging to verify or audit computational integrity \cite{fang2023proof}. As a result, detecting and attributing issues such as trojans or poisoning becomes significantly harder. Therefore, developing comprehensive mechanisms to address privacy and security concerns in this new paradigm is a valuable and important direction for future research.

\section{Related Work}
Studies of FL with LLM are advancing rapidly, attracting numerous attentions and opportunities. We further discuss concurrent related surveys concerning the FedLLM topic \cite{ijcai2024p919,yu2023federated,zhuang2023foundation,chen2024integration,wang2024fedmeki,han2024fedsecurity,yan2025federated}. However, none of these works focus on the evolution of the FedLLM from the white-box to the black-box setting as our paper does.

FedFM \cite{yu2023federated} focuses on exploring the benefits and challenges of integrating FL into the lifespan of FMs, including the pre-training, fine-tuning, and application stages. The survey \cite{zhuang2023foundation} further investigates the synergistic relationship between FL and FMs, considering both the empowerment of FM through FL and the enhancement of FL through FM. Recently, \cite{chen2024integration} explores the potential applications of FedLLM in sectors such as healthcare, finance, and education, aiming to address real-world industry challenges. \cite{wang2024fedmeki} dives into various medical domain tasks of FedLLM. Meanwhile, \cite{ijcai2024p919} adopts a different approach by examining the efficiency of this integration, focusing on two key aspects: computational and communication efficiency. \cite{han2024fedsecurity} focuses on benchmarking the attack and defense approaches on FedLLM. \cite{yan2025federated} compares the different FedLLM fine-tuning strategies that integrate knowledge distillation and split learning.

To bridge the gaps in existing literature, this survey proposes a novel taxonomy that delineates the evolutionary trajectory of federated post-training LLMs, categorized into white-box, gray-box, and black-box settings. Furthermore, we highlight the emerging trend of the inference-only black-box setting for FedLLM.

\section{Conclusions}
Federated large language models (FedLLM) have exhibited significant capabilities in benefiting local users, with recent studies further enhancing their efficiency and privacy protection abilities. In this survey, we introduce a novel taxonomy with two key dimensions to categorize federated tuning LLM approaches: from model accessibility and parameter efficiency. Beyond summarizing the current state of research corresponding to our taxonomy, we highlight the tendency and growing importance of inference-only black-box settings in FedLLM for the future directions.
Additionally, we provide insights into promising future opportunities that we believe hold significant potential and could be substantially advanced through joint development with efficient inference-only black-box tuning.

\bibliography{custom}

\end{document}

%% file: sections/Taxonomy.tex
\section{Federated LLM}
Recently, LLMs have garnered significant attention in both academia and industry, revolutionizing the machine learning paradigm. These models are characterized by an immense parameter size, often scaling to hundreds of billions, and are pre-trained on massive datasets containing up to (tens of) trillions of tokens. The scale of both the models and datasets far exceeds those of previous deep learning applications and what has been explored in federated learning \cite{daly2024federated}.

These LLMs excel as few-shot and zero-shot learners, capable of accomplishing a wide range of tasks through instruction tuning \cite{ye2024openfedllm}, often surpassing smaller, domain-specific models. The success of LLMs heavily depends on access to extensive, high-quality user data, underscoring the growing need for privacy-preserving techniques, particularly FedLLM, in the fine-tuning process.

\subsection{Federated Instruction Tuning}
Post-training, popularized through instruction tuning that integrates supervised fine-tuning with value alignment techniques, has become a standard streamline for tuning LLMs.

\noindent\textbf{Federated Supervised Fine-Tuning.} Conventional supervised fine-tuning (SFT) often serves as the foundation for federated instruction tuning. In this approach, each data sample consists of an instruction-response pair, and the LLM is trained to predict the response based on the given instruction. By unifying a large number of clients within the FL system, the LLM can be trained to effectively follow a diverse range of human instructions.

\noindent\textbf{Federated Value Alignment.} Simply following instructions to generate responses is insufficient, as FedLLM trained through federated SFT may still produce low-quality or even harmful outputs. To address this, federated value alignment is often necessary to ensure that human values are effectively incorporated into FedLLM. Reinforcement learning (RL) methods are commonly employed during the value alignment stage, with popular approaches including direct preference optimization (DPO) and reinforcement learning from human feedback (RLHF).

\subsection{FedLLM Evolving Taxonomy}
Traditionally, knowledge representation bridging the gap between LLM pre-training and downstream tasks is compressed to fit the entire model, following the full model fine-tuning approach. However, recent research indicates that this knowledge does not need to reside entirely within the model. Instead, it can be stored outside the frozen model using additional trainable parameters, as seen in methods like FedLoRA, or associated with input sequences through techniques like FedPrompt. These perspectives explore new ways to make FedLLM more scalable and efficient for collaborative tuning under resource constraints. Experimental results show that these techniques usually achieve effects comparable to (not significantly less than) conventional full model fine-tuning. Therefore, in this paper, we classify these advancements as one axis of FedLLM evolution, referred to as efficiency-based optimization.

Unlike existing surveys that focus solely on efficiency-based optimization, we introduce a novel taxonomy axis for the evolution of FedLLM, termed access-based optimization. We categorize the evolution of FewLLM based on the two axes and show the evolution path in Fig. ~\ref{fig:evolution}.

\begin{figure*}[t]
  \centering  \includegraphics[width=0.92\linewidth]{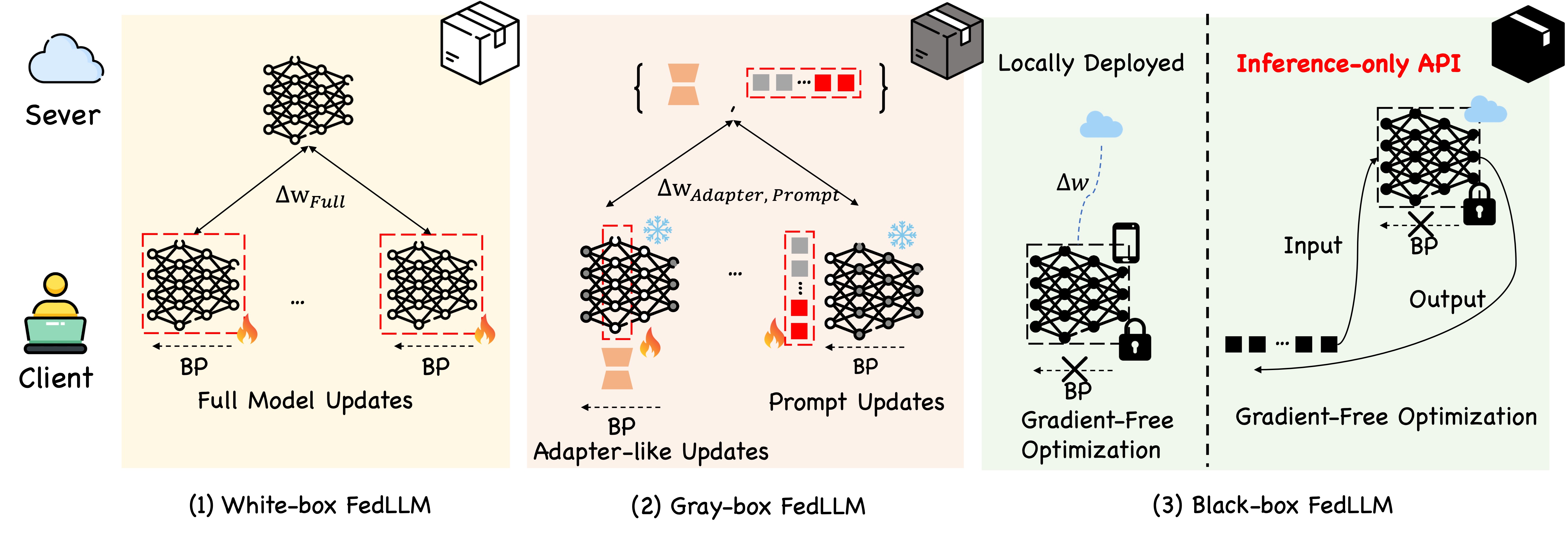}
  \vspace{-0.3cm}
  \caption{Diagram of FedLLM categories. We categorize the existing Federated LLM approaches into the above three types: White-box, Gray-box and Black-box FedLLM. Specifically, white-box employs a full model update and transmission to the server. gray-box employs partial updates and transmit adapter or prompt with the server. black-box leverages a inference-only API, without acquiring the model. Besides, black-box also have another scenario where models are deployed locally without accessing internal information.}
  \vspace{-0.3cm}
  
  \label{fig:FL-framework}
\end{figure*}

\noindent\textbf{Taxonomy Explanation.} As the structure and parameter information of pre-trained LLMs gain increasing commercial value, it has become increasingly challenging for users to gain full access to these models. Instead, they are often limited to updating a partial model or submitting requests via inference APIs. Given this reality, FedLLM must adapt to the settings that support fine-tuning models with partial parameters available or only inference and treat the LLM as a black-box inference API.

Thus, our taxonomy addresses \textit{two} key challenges in Federated LLMs: (1) model accessibility, which includes access to internal information and computation modes of pre-trained LLMs, and (2) parameter efficiency during the optimization process. The taxonomy is illustrated in Figure~\ref{fig:framework}.

\noindent\textbf{Model Accessibility.} The model accessibility dimension denotes the extent of information that can be acquired from the model. We categorize this dimension into `accessible' and `inaccessible'. Specifically, `inaccessible' represents for backpropagation-free mode.

\noindent\textbf{Parameter Efficiency.} The parameter efficiency dimension defines the size of parameters that require updating during training. We categorize it into three levels: full-model update, partial parameters, and input-level model update.

\noindent\textbf{Optimization Method.}
Based on the above two dimensions, we can categorize the tuning methods in Federated LLM into the following three types:1) \textbf{white-box tuning:}
The internal structures and functionalities of the server LLMs are accessible, and the optimization process uses all information and functionalities to update the parameters. 2) \textbf{gray-box tuning:} Some internal information and functionalities of the server LLMs are available, but not completely transparent or completely hidden. Gray-box optimization utilizes partial information and functionalities about the model to guide the optimization process, typically when some model information is accessible. 3) \textbf{black-box tuning:} The internal structure details of the model are inaccessible, and the optimization process relies solely on input-output pairs to guide the search for the optimal solution without using gradient information.

We show the diagram of white-box, gray-box and black-box in Fig.\ref{fig:FL-framework}, and illustrate the corresponding categories respectively in details in the following sections: white-box Sec.\ref{sec:whitebox}, gray-box Sec.\ref{sec:graybox} and black-box Sec.\ref{sec:blackbox}.

%% file: sections/Whitebox.tex
\section{White-Box Tuning on FedLLM}\label{sec:whitebox}
White-box tuning on FedLLM assumes that the server and client LLMs are transparent similar to vanilla federated learning. Here, we revisit the widely-used federated learning algorithm (i.e., FegAvg \cite{mcmahan2017communication}). Suppose that there are $K$ clients in federated learning, and the $k$-th client trains on its local dataset $D_{k}$ with local model parameters $\theta^{k}$. The FedAvg algorithm updates over $T$ communication rounds as follows:
At round \(t = 0\), the server initializes the global model parameters \(\theta^0\), usually by random initialization.
In each round \(t\), the server sends the current global model parameters \(\theta^t\) to all clients \(k = 1,2,\cdots,K\).
After receiving the global model parameters \(\theta^t\), client \(k\) performs \(E\) rounds of training on its local dataset \(D_k\) and obtains the updated local model parameters \(\theta^{k,t + 1}\). This process can be achieved using optimization algorithms such as Stochastic Gradient Descent (SGD). For example:
\(\theta^{k,t + 1}=\theta^{k,t}-\eta\nabla L_{D_k}(\theta^{k,t})\).
Here, \(\eta\) is the learning rate, \(L_{D_k}(\theta^{k,t})\) is the loss function of client \(k\) on the local dataset \(D_k\) with respect to the model parameters \(\theta^{k,t}\), and \(\nabla L_{D_k}(\theta^{k,t})\) is the gradient of the loss function.
After completing local training, each client \(k\) sends the updated local model parameters \(\theta^{k,t + 1}\) back to the server.
Upon receiving updates from all clients, the server aggregates the models and updates the global model parameters \(\theta^{t + 1}\). Usually, simple averaging is used:
\begin{align}
\theta^{t + 1}=\frac{1}{K}\sum_{k = 1}^{K}\theta^{k,t + 1}
\end{align}
If the data size \(n_k\) of each client is different, it is more reasonable to use weighted averaging, and the formula is:
\(\theta^{t + 1}=\sum_{k = 1}^{K}\frac{n_k}{N}\theta^{k,t + 1}\)
where \(N=\sum_{k=1}^{K}n_k\) is the total data size across all clients.

\noindent\textbf{Full Model Fine-Tuning.}
Following FedAvg \cite{mcmahan2017communication}, federated full model fine-tuning of large language models (LLMs) assumes that these models are deployed on both client and server devices, this process we refer to as FedSFT.
Although full model fine-tuning usually outperforms partial model updates \cite{lora}, it is impractical in federated settings due to the immense communication and local training costs, as well as issues related to updated server LLMs ownership \cite{fatellm}. Therefore, in the following sections, we focus on gray-box and black-box tuning on FedLLM from the model access and parameter efficiency perspectives.
\\

%% file: sections/Graybox.tex
\section{Gray-Box Tuning on FedLLM}\label{sec:graybox}


To mitigate the communication and local training issues incurred by federated white-box full model fine-tuning, several parameter-efficient fine-tuning (PEFT) methods have been employed to federated tuning. These methods efficiently update critical internal information and functionality of LLMs. PEFT methods partially update the client in the parameter-efficient manner and then broadcast and aggregate the updated parameters between the client and server, which elegantly alleviates the computation and communication overhead, without requiring full access to the server's full models. Meanwhile, to further preserve the server's ownership of the LLMs, some methods broadcast LLMs proxies to clients, further reducing the accessibility of server LLMs. We define above methods as \textbf{gray-box tuning} because server LLMs are not entirely transparent to clients.
PEFT methods such as low-rank adapters (LoRA) \cite{hu2022lora,NEURIPS2023_3340ee1e,dora}, prompt tuning \cite{lester2021power} have been validated the effectiveness of adapting LLMs to various downstream tasks, and demonstrated less degradation in general ability \cite{lora}. These characteristics are beneficial for tuning LLMs on clients, which often equip low-capability devices and face dynamic tasks. Here, we summarize the PEFT gray-box federated LLM in the \textit{parameter level} (i.e., adapter-based tuning) and \textit{input level} (i.e., prompt tuning), respectively. A brief summary is shown in Table \ref{tab:grayboxcomparison}.

%

\begin{table*}[h]
\centering
{\fontsize{7.5}{10}\selectfont 
\begin{tabular}{llccccc}
\toprule
\textbf{Category}& \textbf{Representative Paper} & \textbf{\begin{tabular}[c]{@{}c@{}}Update\\ Strategy\end{tabular}} & \textbf{\begin{tabular}[c]{@{}c@{}}Communication\\ Efficiency\end{tabular}} & \textbf{\begin{tabular}[c]{@{}c@{}}Training\\ Efficiency\end{tabular}} & \textbf{Accessibility} & \textbf{\begin{tabular}[c]{@{}c@{}}Public\\ Dataset\end{tabular}} \\ 
\midrule
\textbf{\begin{tabular}[l]{@{}l@{}}Full Model\\ Fine-Tuning\end{tabular}} & FedSFT \cite{mcmahan2017communication} & Full Model& Low& Low & High& \xmark\\ 
\midrule
\multirow{5}{*}{\textbf{\begin{tabular}[l]{@{}l@{}}Adapter-like\\ Tuning\end{tabular}}}& FedPepTAO \cite{fedpeptao}& Selected Layers& Medium& Medium& Medium& \xmark \\
& FibecFed \cite{fibecfed} & Selected Layers & Medium & Medium-high  & Medium & \xmark \\
& FedDUMAP \cite{feddumap}  & Selected Layers  & Medium  & Medium  & Medium  & \cmark  \\
& FedOT \cite{fedot} & Adapter  & Medium-high & Medium-high  & Low  & \xmark   \\
& FedBiOT \cite{fedbiot} & Low-rank Adapter & High  & High  & Low  & \cmark                                                               \\ 
\hdashline
\textbf{Prompt Tuning}& Fedprompt \cite{fedprompt}& Prompt& High& High& Medium & \xmark\\ 
\bottomrule
\end{tabular}}
\vspace{-0.2cm}
\caption{Brief summary of the white-box and gray-box tuning on FedLLM. We categorize current work according to their update strategy, communication efficiency, training efficiency, accessibility of public dataset. }
\vspace{-0.15cm}
\label{tab:grayboxcomparison}
\end{table*}

\subsection{Adapter-based gray-box FedLLM}

\textbf{Adapter Tuning.}
FedPepTAO \cite{fedpeptao} dynamically selects a proper set of layers for each device, taking into account the non-Independent and Identically Distributed (non-IID) decentralized data. It only exchanges the prompt parameters of the selected layers during the federated tuning process.
To speed up federated tuning, FibecFed \cite{fibecfed} develops a Fisher information-based device sample pruning method, then adaptively aggregates important layers globally, as indicated by clients' data.
FedDUMAP \cite{feddumap} further incorporates insensitive server public data to speed up convergence and improve accuracy.
HETLORA \cite{HETLORA} argues that naïvely applying
LoRA with homogeneous ranks across clients is sub-optimal and dynamiclly adjusts different
ranks of LoRA to different clients to cater
to the heterogeneous system capabilities and data
complexities of the clients by proposing rank self-pruning and sparsity-weighted aggregation. 
FLoRA \cite{flora} highlights that traditional FL
aggregation strategies on LoRA led to mathematically inaccurate aggregation noise. FLoRA proposes a novel stacking-based aggregation method for aggregating LoRA modules, supporting heterogeneous LoRA ranks across clients.
FedPipe \cite{fedpipe} addresses the heterogeneity of client devices by identifying and prioritizing critical parameters for efficient tuning. It formulates automated federated LLMs PEFT as a Mixed-Integer Linear Programming (MILP) optimization problem and proposes a two-stage method comprising trainable weights identification and a fast search algorithm.

\noindent \textbf{Privacy-preserved Adapter Tuning.}
Recently, the privacy-preserving fine-tuning algorithm known as off-site tuning \cite{offsite} has proven effective by fine-tuning the adapters with the compressed frozen LLM emulator. Specifically, given the LLMs, denoted as $\theta$, offsite-tuning divides $\theta$ into two distinct components: a lightweight and trainable adapter ($\mathcal{A}$), designed for downstream adaptation, and the remaining frozen model ($\varepsilon $), which is compressed into an emulator ($\varepsilon ^{*}$) using the layer-drop-based compression method \cite{layerdrop}. Then, the LLMs \textit{proxy} consisting of both $\mathcal{A}$ and $\varepsilon^{*} $ are sent to the users and the lightweight adapter ($\mathcal{A}$) is optimized for downstream tasks. Only the adapter ($\mathcal{A}$) is sent back to update the original LLMs. Off-site tuning significantly reduces computation overhead while preserving LLMs' ownership in the gray-box setting.
FedOT \cite{fedot} and FATE-LLM \cite{fatellm} extand offsite-tuning to the federated version. The server broadcasts the light-weight LLMs \textit{proxy} ($\mathcal{A}$ and $\varepsilon^{*}$) to all clients, and the clients only fine-tune the adapter ($\mathcal{A}$) following FedAvg \cite{mcmahan2017communication}. FedOT especially alleviates the huge communication overhead of broadcasting LLMs, which is a bottleneck in federated learning. However, FedOT does not address the significant distribution shift between the datasets of the clients and the server. To tackle this issue, FedBiOT \cite{fedbiot} formulates a bi-level optimization to minimize the negative effect of data discrepancy and derive the updating rules for the server and clients, i.e., the clients perform multiple local updates to fine-tune the adapter, and the server distills the emulator with public datasets from the original LLMs, while aggregating the updated adapters from the clients. To further reduce the communication overhead induced by the bi-level optimization, adapter and emulator LoRA \cite{hu2022lora} ($\mathcal{A}_{lora}$ and $\varepsilon ^{*}_{lora}$) are broadcast during the bi-level optimization. The optimization is generally simplified as follows:

{
\vspace{-5pt}
\small
\begin{align}
&\min_{\mathcal{A}_{lora} } \sum_{m \in M}p_{m}F_{m}(\mathcal{A}, \varepsilon ^{*}) 
\label{eq_out}\\  
s.t. \varepsilon ^{*}_{lora} & \in arg \min_{\varepsilon ^{*}_{lora} }\frac{1}{\mathcal{D}} 
{\textstyle \sum_{x \in \mathcal{D}}} \left [||\varepsilon, \varepsilon ^{*}|| + ||\theta, \varepsilon ^{*}\cup\mathcal{A}||\right ] \label{eq_in} 
\end{align}
}
where the inner-level optimization. i.e., Eq. \ref{eq_in}, performs on the server, and it constrains the distance ($||\cdot ||$) of the emulator and the full model as well as model-adapter combination and the emulator-adapter via the public datasets ($\mathcal{D}$). The server then broadcasts the aligned $\varepsilon ^{*}_{lora}$ to clients for further collaborative updating. The outer-level optimization. i.e., Eq. \ref{eq_out}, performs local clients updating then communicates $\mathcal{A}_{lora}$ with the server.
FedPFT \cite{fedpft} argues that layer-drop-based compression in offsite-tuning causes mismatched and insufficient fine-tuning. Therefore, FedPFT proposes a layer-wise compression to emphasizes the crucial neurons across all layers and conducts a layer-level and neuron-level knowledge distillation to accurately align compressed LLMs with the original LLMs.  

\subsection{Prompt tuning gray-box FedLLM}
To further reduce communication costs and restrict accessibility of server LLMs, prompt tuning-like methods \cite{lester2021power,li2021prefix,liu2024gpt}, which tune LLMs' input-level parameters, have garnered significant research interest \cite{fatellm}. Fedprompt \cite{fedprompt} tunes the soft (or trainable) prompts that precede the input text and then aggregates these soft prompts on the server, updating and broadcasting only 0.01$\%$ of the LLMs’ parameters. PTCC \cite{ptcc} analyzes the limited representation capability of existing soft-prompt initialization methods and proposes the task-aware initialization approach, which continually adapts LLMs and mitigates catastrophic forgetting. To enable sufficient personalization of prompt, pFedPrompt \cite{pfedprompt} learns user consensus in linguistic space and adapts to user features on each client in visual space in a non-parametric manner, respectively. CaFPT \cite{cafpt} revisits the prompt tuning and identifies the data-sensitive paradigm of local tuning. Consequently, context-aware prompt tuning strategy is proposed to select the most pertinent client data samples.          

%% file: sections/Blackbox.tex
\section{Black-box Tuning on FedLLM}\label{sec:blackbox}

\begin{table*}[h]
    \centering
    {\fontsize{7.5}{10}\selectfont 
    \begin{tabular}{lccccccccc}
        \toprule
        \multirow{2}*{\textbf{Category}}&\multirow{2}*{\textbf{Representative Paper}} & \textbf{Update } & \textbf{LLM} & \textbf{Optimization} & \textbf{Communication} & \textbf{Data}  \\
        & & \textbf{Strategy } & \textbf{Deployment} & \textbf{Metrics} & \textbf{Efficiency} & \textbf{Heterogeneity}  \\
        \midrule
        &FedMeZO \cite{ling2024convergence} & Full & Local & ZOO & LoRA & \cmark \\
        \textbf{Inference-only}&FedKSeed \cite{qinfederated} & Full & Local & ZOO & Accumulator & \textit{None}  \\
        \textbf{Locally-}&DP-ZO-FedSGD \cite{qinfederated} & Full & Local & ZOO & Real-time Gradients & \textit{None}  \\
        \textbf{Deployed}&FwdLLM \cite{xu2024fwdllm} & PEFT & Local & Perturbation & PEFT & \textit{None}  \\
        &ZOOPFL \cite{luzoopfl} & Prompt & Local & ZOO & Auto-encoder & \cmark  \\
        &Wang et al. \cite{wang2024personalized} & Prompt & Local &  Disc. Local Search & Disc. Tokens & \cmark   \\
        \hdashline
        &FedBPT \cite{sunfedbpt} & Prompt & API & CMA-ES & CMA-ES Para& \textit{None}  \\
        \textbf{Inference-only}&Li et al. \cite{li2024federated} & Prompt & API & CMA-ES & HTTP-based   & \textit{None} \\
        \textbf{APIs}&FedDTPT \cite{wu2024feddtpt} & Prompt & API & Feedback Loop  & Discrete Tokens &\cmark \\
        &Fed-BBPT \cite{lin2023efficient} & Prompt & API & ZOO & Generator & \cmark  \\
        \bottomrule
    \end{tabular}}
    \vspace{-0.2cm}
    \caption{Summary of the black-box tuning on FedLLM. We categorize current work according to their update strategy, LLM deployed mode, optimization metrics, communication efficiency and date heterogeneity. \textit{None} denotes the particular element is not specifically targeted for this work.}
    \vspace{-0.15cm}
    \label{tab:blackboxcomparison}
\end{table*}

Although federated parameter-efficient methods have been successful in enhancing the performance of LLMs over recent years, significant challenges persist in accessing internal model information. This situation is even more challenging for local users, especially with the increasing privacy concerns and the growing emphasis on data protection in contemporary society.
Consequently, to satisfy both the privacy demand along with the substantial requirements of using LLMs on client, the concept of LLMs as an API service has emerged. Yet, leveraging manually crafted text prompts for particular downstream tasks always leads to unsatisfactory performance in most instances. A brief summary is shown in Table \ref{tab:blackboxcomparison}.

\noindent\textbf{Black-box Tuning for LLMs-as-a-Service (LMaaS).}
Therefore, it is crucial to develop an approach to enhance the API performance for local user tasks, aiming to leverage black-box tuning to optimize the continuous prompt even in the absence of backward propagation optimization.
An early exploration of the endeavor is BBT \cite{sun2022black}, which projects the prompt space onto a smaller subspace using random linear projection, solving the optimization with derivative-free methods. To enhance the versatility of BBT, BBTv2 \cite{sun2022bbtv2} prepends prompts to hidden states at every layer of the PTM and optimizes the prompts at different layers alternately. Unlike the previous effort on continuous prompt optimization, BDPL \cite{diaoblack} tackles discrete prompts optimization by applying a variance-reduced policy gradient algorithm to estimate the gradient of the categorical distribution and update the discrete prompts respectively. To alleviate substantial number of API calls due to the random choice of the subspace, \cite{zhang2024subspace} utilize a zeroth-order mini-batch stochastic proximal gradient method to accelerate the optimal subspaces selection process for prompt optimization.

\textbf{Federated Gradient-free Optimization with Full Model.}
In parallel, federated learning as a solution in collaboratively fine-tuning pre-trained LLMs on decentralized data while maintaining data privacy is gaining widespread attention. Researches in Sec.\ref{sec:graybox} put efforts into minimizing memory usage and communication overhead, simultaneously limiting direct access to the LLMs. Although reducing the dependence on the model, PEFT-based methods still demand side information for adaptation like model architecture or weights for optimization. In order to remove the complete reliance on model gradients, FedMeZO \cite{ling2024convergence} incorporating a memory-efficient Zeroth-Order Optimization method, MeZO \cite{malladi2023fine} into FL and analyzes the convergence to examine the theoretical underpinnings of ZOO-based FL in the context of LLMs. FedMeZO further leverages LoRA to mitigate the high communication cost.Later on, FedKSeed \cite{qinfederated} improves ZOO-based methods by employing with a finite set of random seeds to facilitate a better communication efficiency. Furthermore, DP-ZO-FedSGD \cite{xing2025privacy} proposes a DP-based FL algorithm using the ZO oracles, enabling efficient and privacy-preserving full-parameter fine-tuning without relying on historical client gradients.

\noindent\textbf{Parameter-Efficient Federated Gradient-free Tuning.}
However, federated learning faces several challenges with black-box optimization in the full model setting, as the significant computation and communication overhead intensify especially when the model parameters increase in size. Therefore, FwdLLM \cite{xu2024fwdllm} integrates the perturbed inferences with PEFT techniques like LoRA and Adapter on each device to deliver better memory efficiency as well as faster convergence speed. Other method like ZOOPFL \cite{luzoopfl} brings an auto-encoder with low embedding dimensions to represent transformations and learns the transformations on the inputs and mappings of the outputs through zeroth-order optimization.

\noindent\textbf{Efficient Federated Black-box Prompt Tuning with Inference-only APIs.}
Furthermore, researchers employ LLMs-as-a-Service (LMaaS) along with federated learning to facilitate efficient and privacy-preserving prompt tuning across decentralized datasets. As an early exploration, FedBPT \cite{sunfedbpt} adopts a gradient-free optimization that leverages CMA-ES to search for optimal distributions of the continuous prompt based on local data. Later, \cite{li2024federated} builds a real-world implementation of the FedBPT system across various edge devices.
Unlike FedBPT, FedDTPT \cite{wu2024feddtpt} proposes a token-level discrete prompt optimization method, which leverages a new feedback mechanism for inference results to drive gradient-free prompt optimization. Similarly, \cite{wang2024personalized} focuses on the discrete tokens with discrete local search, while further reducing the memory and communication cost by compressing the prompt embeddings with discrete tokens. Other method like Fed-BBPT \cite{lin2023efficient} introduces a lightweight prompt generator
tailored to their local dataset on each local user and adpots ZOO technique to evaluate the contribution of the generated prompts for the local task.

%% file: main.bbl
\begin{thebibliography}{54}
\providecommand{\natexlab}[1]{#1}

\bibitem[{Bai et~al.(2022{\natexlab{a}})Bai, Jones, Ndousse, Askell, Chen, DasSarma, Drain, Fort, Ganguli, Henighan, Joseph, Kadavath, Kernion, Conerly, El-Showk, Elhage, Hatfield-Dodds, Hernandez, Hume, Johnston, Kravec, Lovitt, Nanda, Olsson, Amodei, Brown, Clark, McCandlish, Olah, Mann, and Kaplan}]{bai2022training}
Yuntao Bai, Andy Jones, Kamal Ndousse, Amanda Askell, Anna Chen, Nova DasSarma, Dawn Drain, Stanislav Fort, Deep Ganguli, Tom Henighan, Nicholas Joseph, Saurav Kadavath, Jackson Kernion, Tom Conerly, Sheer El-Showk, Nelson Elhage, Zac Hatfield-Dodds, Danny Hernandez, Tristan Hume, and 12 others. 2022{\natexlab{a}}.
\newblock Training a helpful and harmless assistant with reinforcement learning from human feedback.
\newblock \emph{arXiv preprint arXiv:2204.05862}.

\bibitem[{Bai et~al.(2022{\natexlab{b}})Bai, Kadavath, Kundu, and et~al.}]{bai2022constitutional}
Yuntao Bai, Saurav Kadavath, Sandipan Kundu, and et~al. 2022{\natexlab{b}}.
\newblock Constitutional ai: Harmlessness from ai feedback.
\newblock \emph{arXiv preprint arXiv:2212.08073}.

\bibitem[{Biderman et~al.(2024)Biderman, Portes, Ortiz, Paul, Greengard, Jennings, King, Havens, Chiley, Frankle et~al.}]{lora}
Dan Biderman, Jacob Portes, Jose Javier~Gonzalez Ortiz, Mansheej Paul, Philip Greengard, Connor Jennings, Daniel King, Sam Havens, Vitaliy Chiley, Jonathan Frankle, and 1 others. 2024.
\newblock Lora learns less and forgets less.
\newblock \emph{arXiv preprint arXiv:2405.09673}.

\bibitem[{Che et~al.(2023)Che, Liu, Zhou, Ren, Zhou, Sheng, Dai, and Dou}]{fedpeptao}
Tianshi Che, Ji~Liu, Yang Zhou, Jiaxiang Ren, Jiwen Zhou, Victor~S Sheng, Huaiyu Dai, and Dejing Dou. 2023.
\newblock Federated learning of large language models with parameter-efficient prompt tuning and adaptive optimization.
\newblock \emph{Conference on Empirical Methods in Natural Language Processing}.

\bibitem[{Chen et~al.(2024)Chen, Feng, Li, Lyu, Zhou, Zheng, and Yin}]{chen2024integration}
Chaochao Chen, Xiaohua Feng, Yuyuan Li, Lingjuan Lyu, Jun Zhou, Xiaolin Zheng, and Jianwei Yin. 2024.
\newblock Integration of large language models and federated learning.
\newblock 5(12).

\bibitem[{Cho et~al.(2024)Cho, Liu, Xu, Fahrezi, and Joshi}]{HETLORA}
Yae~Jee Cho, Luyang Liu, Zheng Xu, Aldi Fahrezi, and Gauri Joshi. 2024.
\newblock Heterogeneous lora for federated fine-tuning of on-device foundation models.
\newblock \emph{arXiv preprint arXiv:2401.06432}.

\bibitem[{Daly et~al.(2024)Daly, Eichner, Kairouz, McMahan, Ramage, and Xu}]{daly2024federated}
Katharine Daly, Hubert Eichner, Peter Kairouz, H~Brendan McMahan, Daniel Ramage, and Zheng Xu. 2024.
\newblock Federated learning in practice: Reflections and projections.
\newblock \emph{arXiv preprint arXiv:2410.08892}.

\bibitem[{Diao et~al.()Diao, Huang, Xu, Li, Lin, Zhou, and Zhang}]{diaoblack}
Shizhe Diao, Zhichao Huang, Ruijia Xu, Xuechun Li, Yong Lin, Xiao Zhou, and Tong Zhang.
\newblock Black-box prompt learning for pre-trained language models.
\newblock \emph{Transactions on Machine Learning Research}.

\bibitem[{Fan et~al.(2023)Fan, Kang, Ma, Chen, Wei, Fan, and Yang}]{fatellm}
Tao Fan, Yan Kang, Guoqiang Ma, Weijing Chen, Wenbin Wei, Lixin Fan, and Qiang Yang. 2023.
\newblock Fate-llm: A industrial grade federated learning framework for large language models.
\newblock \emph{arXiv preprint arXiv:2310.10049}.

\bibitem[{Fang et~al.(2023)Fang, Jia, Thudi, Yaghini, Choquette-Choo, Dullerud, Chandrasekaran, and Papernot}]{fang2023proof}
Congyu Fang, Hengrui Jia, Anvith Thudi, Mohammad Yaghini, Christopher~A Choquette-Choo, Natalie Dullerud, Varun Chandrasekaran, and Nicolas Papernot. 2023.
\newblock Proof-of-learning is currently more broken than you think.
\newblock In \emph{2023 IEEE 8th European Symposium on Security and Privacy (EuroS\&P)}, pages 797--816. IEEE.

\bibitem[{Fang et~al.(2024)Fang, Lin, Chen, Chen, Gao, and Fang}]{fedpipe}
Zihan Fang, Zheng Lin, Zhe Chen, Xianhao Chen, Yue Gao, and Yuguang Fang. 2024.
\newblock Automated federated pipeline for parameter-efficient fine-tuning of large language models.
\newblock \emph{arXiv preprint arXiv:2404.06448}.

\bibitem[{Gu et~al.(2023)Gu, Wang, Wu, and et~al.}]{NEURIPS2023_3340ee1e}
Yuchao Gu, Xintao Wang, Jay~Zhangjie Wu, and et~al. 2023.
\newblock Mix-of-show: Decentralized low-rank adaptation for multi-concept customization of diffusion models.
\newblock In \emph{Advances in Neural Information Processing Systems}, volume~36, pages 15890--15902. Curran Associates, Inc.

\bibitem[{Guo et~al.(2023)Guo, Guo, and Wang}]{pfedprompt}
Tao Guo, Song Guo, and Junxiao Wang. 2023.
\newblock pfedprompt: Learning personalized prompt for vision-language models in federated learning.
\newblock In \emph{Proceedings of the ACM Web Conference}, page 1364–1374.

\bibitem[{Guo et~al.(2024)Guo, Guo, and Wang}]{cafpt}
Tao Guo, Song Guo, and Junxiao Wang. 2024.
\newblock Explore and cure: Unveiling sample effectiveness with context-aware federated prompt tuning.
\newblock \emph{IEEE Transactions on Mobile Computing}.

\bibitem[{Han et~al.(2024)Han, Buyukates, Hu, Jin, Jin, Sun, Wang, Wu, Xie, Yao, Zhang, Zhang, Zhang, Joe-Wong, Avestimehr, and He}]{han2024fedsecurity}
Shanshan Han, Baturalp Buyukates, Zijian Hu, Han Jin, Weizhao Jin, Lichao Sun, Xiaoyang Wang, Wenxuan Wu, Chulin Xie, Yuhang Yao, Kai Zhang, Qifan Zhang, Yuhui Zhang, Carlee Joe-Wong, Salman Avestimehr, and Chaoyang He. 2024.
\newblock Fedsecurity: A benchmark for attacks and defenses in federated learning and federated llms.
\newblock In \emph{ACM SIGKDD Conference on Knowledge Discovery and Data Mining}, pages 5070--5081.

\bibitem[{Hu et~al.(2022)Hu, Shen, Wallis, Allen-Zhu, Li, Wang, Wang, Chen et~al.}]{hu2022lora}
Edward~J Hu, Yelong Shen, Phillip Wallis, Zeyuan Allen-Zhu, Yuanzhi Li, Shean Wang, Lu~Wang, Weizhu Chen, and 1 others. 2022.
\newblock Lora: Low-rank adaptation of large language models.
\newblock In \emph{International Conference on Learning Representations}.

\bibitem[{Kuang et~al.(2024)Kuang, Qian, Li, Chen, Gao, Pan, Xie, Li, Ding, and Zhou}]{fedot}
Weirui Kuang, Bingchen Qian, Zitao Li, Daoyuan Chen, Dawei Gao, Xuchen Pan, Yuexiang Xie, Yaliang Li, Bolin Ding, and Jingren Zhou. 2024.
\newblock Federatedscope-llm: A comprehensive package for fine-tuning large language models in federated learning.
\newblock In \emph{ACM SIGKDD Conference on Knowledge Discovery and Data Mining}, page 5260–5271.

\bibitem[{Lee et~al.(2024)Lee, Phatale, Mansoor, and et~al.}]{lee24rlaif}
Harrison Lee, Samrat Phatale, Hassan Mansoor, and et~al. 2024.
\newblock Rlaif vs. rlhf: Scaling reinforcement learning from human feedback with ai feedback.
\newblock In \emph{International Conference on Machine Learning}.

\bibitem[{Lester et~al.(2021)Lester, Al-Rfou, and Constant}]{lester2021power}
Brian Lester, Rami Al-Rfou, and Noah Constant. 2021.
\newblock The power of scale for parameter-efficient prompt tuning.
\newblock In \emph{Conference on Empirical Methods in Natural Language Processing}, pages 3045--3059.

\bibitem[{Li and Liang(2021)}]{li2021prefix}
Xiang~Lisa Li and Percy Liang. 2021.
\newblock Prefix-tuning: Optimizing continuous prompts for generation.
\newblock \emph{arXiv preprint arXiv:2101.00190}.

\bibitem[{Li et~al.(2024)Li, Sun, Liu, Zhang, Li, Chen, Roth, Xu, Chen, and Chen}]{li2024federated}
Yiming Li, Jingwei Sun, Yudong Liu, Yuandong Zhang, Ang Li, Beidi Chen, Holger~R Roth, Daguang Xu, Tingjun Chen, and Yiran Chen. 2024.
\newblock Federated black-box prompt tuning system for large language models on the edge.
\newblock In \emph{ACM International Conference on Mobile Computing and Networking}, pages 1775--1777.

\bibitem[{Lin et~al.(2023)Lin, Sun, Shi, Wang, Huang, Shen, and Tao}]{lin2023efficient}
Zihao Lin, Yan Sun, Yifan Shi, Xueqian Wang, Lifu Huang, Li~Shen, and Dacheng Tao. 2023.
\newblock Efficient federated prompt tuning for black-box large pre-trained models.
\newblock \emph{arXiv preprint arXiv:2310.03123}.

\bibitem[{Ling et~al.(2024)Ling, Chen, Yao, Li, and Shen}]{ling2024convergence}
Zhenqing Ling, Daoyuan Chen, Liuyi Yao, Yaliang Li, and Ying Shen. 2024.
\newblock On the convergence of zeroth-order federated tuning for large language models.
\newblock In \emph{ACM SIGKDD Conference on Knowledge Discovery and Data Mining}, pages 1827--1838.

\bibitem[{Liu et~al.(2024{\natexlab{a}})Liu, Jia, Zhang, Yun, Wang, Zhou, Dai, and Dou}]{feddumap}
Ji~Liu, Juncheng Jia, Hong Zhang, Yuhui Yun, Leye Wang, Yang Zhou, Huaiyu Dai, and Dejing Dou. 2024{\natexlab{a}}.
\newblock Efficient federated learning using dynamic update and adaptive pruning with momentum on shared server data.
\newblock \emph{ACM Transactions on Intelligent Systems and Technology}.

\bibitem[{Liu et~al.(2024{\natexlab{b}})Liu, Ren, Jin, Zhang, Zhou, Valduriez, and Dou}]{fibecfed}
Ji~Liu, Jiaxiang Ren, Ruoming Jin, Zijie Zhang, Yang Zhou, Patrick Valduriez, and Dejing Dou. 2024{\natexlab{b}}.
\newblock Fisher information-based efficient curriculum federated learning with large language models.
\newblock \emph{Conference on Empirical Methods in Natural Language Processing}.

\bibitem[{Liu et~al.(2024{\natexlab{c}})Liu, Zheng, Du, Ding, Qian, Yang, and Tang}]{liu2024gpt}
Xiao Liu, Yanan Zheng, Zhengxiao Du, Ming Ding, Yujie Qian, Zhilin Yang, and Jie Tang. 2024{\natexlab{c}}.
\newblock Gpt understands, too.
\newblock \emph{AI Open}, 5:208--215.

\bibitem[{Lu et~al.(2023)Lu, Yu, Wang, Teney, Wang, Zhu, Chen, Yang, Xie, and Ji}]{luzoopfl}
Wang Lu, Hao Yu, Jindong Wang, Damien Teney, Haohan Wang, Yao Zhu, Yiqiang Chen, Qiang Yang, Xing Xie, and Xiangyang Ji. 2023.
\newblock Zoopfl: Exploring black-box foundation models for personalized federated learning.
\newblock \emph{arXiv preprint arXiv:2310.05143}.

\bibitem[{Malladi et~al.(2023)Malladi, Gao, Nichani, Damian, Lee, Chen, and Arora}]{malladi2023fine}
Sadhika Malladi, Tianyu Gao, Eshaan Nichani, Alex Damian, Jason~D Lee, Danqi Chen, and Sanjeev Arora. 2023.
\newblock Fine-tuning language models with just forward passes.
\newblock \emph{Advances in Neural Information Processing Systems}, 36:53038--53075.

\bibitem[{McMahan et~al.(2017)McMahan, Moore, Ramage, Hampson, and y~Arcas}]{mcmahan2017communication}
Brendan McMahan, Eider Moore, Daniel Ramage, Seth Hampson, and Blaise~Aguera y~Arcas. 2017.
\newblock Communication-efficient learning of deep networks from decentralized data.
\newblock In \emph{Artificial intelligence and statistics}, pages 1273--1282. PMLR.

\bibitem[{Peng et~al.(2024)Peng, Fan, Chen, Wang, Pan, Wen, Zhang, and Wang}]{fedpft}
Zhaopeng Peng, Xiaoliang Fan, Yufan Chen, Zheng Wang, Shirui Pan, Chenglu Wen, Ruisheng Zhang, and Cheng Wang. 2024.
\newblock Fedpft: Federated proxy fine-tuning of foundation models.
\newblock \emph{International Joint Conference on Artificial Intelligence}.

\bibitem[{Qin et~al.(2024)Qin, Chen, Qian, Ding, Li, and Deng}]{qinfederated}
Zhen Qin, Daoyuan Chen, Bingchen Qian, Bolin Ding, Yaliang Li, and Shuiguang Deng. 2024.
\newblock Federated full-parameter tuning of billion-sized language models with communication cost under 18 kilobytes.
\newblock In \emph{International Conference on Machine Learning}.

\bibitem[{Rafailov et~al.(2023)Rafailov, Sharma, Mitchell, Manning, Ermon, and Finn}]{rafailov2023direct}
Rafael Rafailov, Archit Sharma, Eric Mitchell, Christopher~D Manning, Stefano Ermon, and Chelsea Finn. 2023.
\newblock Direct preference optimization: Your language model is secretly a reward model.
\newblock \emph{Advances in Neural Information Processing Systems}, 36:53728--53741.

\bibitem[{Sajjad et~al.(2023)Sajjad, Dalvi, Durrani, and Nakov}]{layerdrop}
Hassan Sajjad, Fahim Dalvi, Nadir Durrani, and Preslav Nakov. 2023.
\newblock On the effect of dropping layers of pre-trained transformer models.
\newblock \emph{Computer Speech \& Language}, 77:101429.

\bibitem[{Sun et~al.(2025)Sun, Xu, Yin, Yang, Xu, Chen, and Roth}]{sunfedbpt}
Jingwei Sun, Ziyue Xu, Hongxu Yin, Dong Yang, Daguang Xu, Yiran Chen, and Holger~R. Roth. 2025.
\newblock Fedbpt: Efficient federated black-box prompt tuning for large language models.
\newblock In \emph{International Conference on Machine Learning}.

\bibitem[{Sun et~al.(2022{\natexlab{a}})Sun, He, Qian, Zhou, Huang, and Qiu}]{sun2022bbtv2}
Tianxiang Sun, Zhengfu He, Hong Qian, Yunhua Zhou, Xuanjing Huang, and Xipeng Qiu. 2022{\natexlab{a}}.
\newblock Bbtv2: Towards a gradient-free future with large language models.
\newblock In \emph{Conference on Empirical Methods in Natural Language Processing}, pages 3916--3930.

\bibitem[{Sun et~al.(2022{\natexlab{b}})Sun, Shao, Qian, Huang, and Qiu}]{sun2022black}
Tianxiang Sun, Yunfan Shao, Hong Qian, Xuanjing Huang, and Xipeng Qiu. 2022{\natexlab{b}}.
\newblock Black-box tuning for language-model-as-a-service.
\newblock In \emph{International Conference on Machine Learning}, pages 20841--20855.

\bibitem[{Wang et~al.(2024{\natexlab{a}})Wang, Wang, Lyu, Chen, and Ma}]{wang2024fedmeki}
Jiaqi Wang, Xiaochen Wang, Lingjuan Lyu, Jinghui Chen, and Fenglong Ma. 2024{\natexlab{a}}.
\newblock {FEDMEKI}: A benchmark for scaling medical foundation models via federated knowledge injection.
\newblock In \emph{The Thirty-eight Conference on Neural Information Processing Systems Datasets and Benchmarks Track}.

\bibitem[{Wang et~al.()Wang, Yu, Zhang, Kim, Rossi, Zhao, Wu, Mitra, Yao, and Henao}]{wang2024personalized}
Rui Wang, Tong Yu, Ruiyi Zhang, Sungchul Kim, Ryan Rossi, Handong Zhao, Junda Wu, Subrata Mitra, Lina Yao, and Ricardo Henao.
\newblock Personalized federated learning for text classification with gradient-free prompt tuning.
\newblock In \emph{Findings of the Association for Computational Linguistics: NAACL 2024}, pages 4597--4612.

\bibitem[{Wang et~al.(2024{\natexlab{b}})Wang, Shen, He, Sun, Wang, Lyu, and Li}]{flora}
Ziyao Wang, Zheyu Shen, Yexiao He, Guoheng Sun, Hongyi Wang, Lingjuan Lyu, and Ang Li. 2024{\natexlab{b}}.
\newblock Flora: Federated fine-tuning large language models with heterogeneous low-rank adaptations.
\newblock \emph{arXiv preprint arXiv:2409.05976}.

\bibitem[{Woisetschl{\"a}ger et~al.(2024)Woisetschl{\"a}ger, Isenko, Wang, Mayer, and Jacobsen}]{ijcai2024p919}
Herbert Woisetschl{\"a}ger, Alexander Isenko, Shiqiang Wang, Ruben Mayer, and Hans-Arno Jacobsen. 2024.
\newblock A survey on efficient federated learning methods for foundation model training.
\newblock In \emph{International Joint Conference on Artificial Intelligence}, pages 8317--8325.

\bibitem[{Wu et~al.(2024{\natexlab{a}})Wu, Li, Li, Ding, and Gao}]{fedbiot}
Feijie Wu, Zitao Li, Yaliang Li, Bolin Ding, and Jing Gao. 2024{\natexlab{a}}.
\newblock Fedbiot: Llm local fine-tuning in federated learning without full model.
\newblock In \emph{ACM SIGKDD Conference on Knowledge Discovery and Data Mining}, page 3345–3355.

\bibitem[{Wu et~al.(2024{\natexlab{b}})Wu, Chen, Yang, Li, Hou, Jing, Wang, Chen, and Tian}]{wu2024feddtpt}
Jiaqi Wu, Simin Chen, Yuzhe Yang, Yijiang Li, Shiyue Hou, Rui Jing, Zehua Wang, Wei Chen, and Zijian Tian. 2024{\natexlab{b}}.
\newblock Feddtpt: Federated discrete and transferable prompt tuning for black-box large language models.
\newblock \emph{arXiv preprint arXiv:2411.00985}.

\bibitem[{Xiao et~al.(2023)Xiao, Lin, and Han}]{offsite}
Guangxuan Xiao, Ji~Lin, and Song Han. 2023.
\newblock Offsite-tuning: Transfer learning without full model.
\newblock \emph{arXiv preprint arXiv:2302.04870}.

\bibitem[{Xing et~al.(2025)Xing, Dong, Hu, Leung, Deen, and Guo}]{xing2025privacy}
Ke~Xing, Yanjie Dong, Xiping Hu, Victor~CM Leung, M~Jamal Deen, and Song Guo. 2025.
\newblock Privacy-aware federated fine-tuning of large pretrained models with just forward propagation.
\newblock In \emph{ICASSP 2025-2025 IEEE International Conference on Acoustics, Speech and Signal Processing (ICASSP)}, pages 1--5. IEEE.

\bibitem[{Xu et~al.(2024)Xu, Cai, Wu, Li, and Wang}]{xu2024fwdllm}
Mengwei Xu, Dongqi Cai, Yaozong Wu, Xiang Li, and Shangguang Wang. 2024.
\newblock Fwdllm: Efficient federated finetuning of large language models with perturbed inferences.
\newblock In \emph{USENIX Conference on Usenix Annual Technical Conference}, pages 579--596.

\bibitem[{Yan et~al.(2025)Yan, Su, Deng, and Schober}]{yan2025federated}
Na~Yan, Yang Su, Yansha Deng, and Robert Schober. 2025.
\newblock Federated fine-tuning of llms: Framework comparison and research directions.
\newblock \emph{arXiv preprint arXiv:2501.04436}.

\bibitem[{yang Liu et~al.(2024)yang Liu, Wang, Yin, Molchanov, Wang, Cheng, and Chen}]{dora}
Shih yang Liu, Chien-Yi Wang, Hongxu Yin, Pavlo Molchanov, Yu-Chiang~Frank Wang, Kwang-Ting Cheng, and Min-Hung Chen. 2024.
\newblock \href {https://openreview.net/forum?id=3d5CIRG1n2} {Do{RA}: Weight-decomposed low-rank adaptation}.
\newblock In \emph{Forty-first International Conference on Machine Learning}.

\bibitem[{Yao et~al.(2024)Yao, Zhang, Wu, Huang, Xia, Yu, Zhang, Kim, Rossi, Li, Yao, McAuley, Chen, and Joe-Wong}]{yao2024federated}
Yuhang Yao, Jianyi Zhang, Junda Wu, Chengkai Huang, Yu~Xia, Tong Yu, Ruiyi Zhang, Sungchul Kim, Ryan Rossi, Ang Li, Lina Yao, Julian McAuley, Yiran Chen, and Carlee Joe-Wong. 2024.
\newblock Federated large language models: Current progress and future directions.
\newblock \emph{arXiv preprint arXiv:2409.15723}.

\bibitem[{Ye et~al.(2024)Ye, Wang, Chai, Li, Li, Xu, Du, Wang, and Chen}]{ye2024openfedllm}
Rui Ye, Wenhao Wang, Jingyi Chai, Dihan Li, Zexi Li, Yinda Xu, Yaxin Du, Yanfeng Wang, and Siheng Chen. 2024.
\newblock Openfedllm: Training large language models on decentralized private data via federated learning.
\newblock In \emph{ACM SIGKDD Conference on Knowledge Discovery and Data Mining}, pages 6137--6147.

\bibitem[{Yu et~al.(2023)Yu, Mu{\~n}oz, and Jannesari}]{yu2023federated}
Sixing Yu, J~Pablo Mu{\~n}oz, and Ali Jannesari. 2023.
\newblock Federated foundation models: Privacy-preserving and collaborative learning for large models.
\newblock \emph{arXiv preprint arXiv:2305.11414}.

\bibitem[{Zhang et~al.(2024{\natexlab{a}})Zhang, Zhang, Gu, and Chang}]{zhang2024subspace}
Haozhen Zhang, Hualin Zhang, Bin Gu, and Yi~Chang. 2024{\natexlab{a}}.
\newblock Subspace selection based prompt tuning with nonconvex nonsmooth black-box optimization.
\newblock In \emph{ACM SIGKDD Conference on Knowledge Discovery and Data Mining}, pages 4179--4190.

\bibitem[{Zhang et~al.(2024{\natexlab{b}})Zhang, Yu, Zhao, Xie, Yao, and Li}]{ptcc}
Zhehao Zhang, Tong Yu, Handong Zhao, Kaige Xie, Lina Yao, and Shuai Li. 2024{\natexlab{b}}.
\newblock Exploring soft prompt initialization strategy for few-shot continual text classification.
\newblock In \emph{International Conference on Acoustics, Speech, and Signal Processing}, pages 12106--12110. IEEE.

\bibitem[{Zhao et~al.(2023)Zhao, Du, Li, Li, and Liu}]{fedprompt}
Haodong Zhao, Wei Du, Fangqi Li, Peixuan Li, and Gongshen Liu. 2023.
\newblock Fedprompt: Communication-efficient and privacy-preserving prompt tuning in federated learning.
\newblock In \emph{International Conference on Acoustics, Speech, and Signal Processing}, pages 1--5. IEEE.

\bibitem[{Zhuang et~al.(2023)Zhuang, Chen, and Lyu}]{zhuang2023foundation}
Weiming Zhuang, Chen Chen, and Lingjuan Lyu. 2023.
\newblock When foundation model meets federated learning: Motivations, challenges, and future directions.
\newblock \emph{arXiv preprint arXiv:2306.15546}.

\end{thebibliography}
